\documentclass[10pt,twocolumn,letterpaper]{article}

\usepackage[]{cvpr}
\usepackage[pagebackref,breaklinks,colorlinks,citecolor=blue]{hyperref}
\usepackage{indentfirst} 
\usepackage{times}
\usepackage{epsfig}
\usepackage{graphicx}
\usepackage{amsmath}
\usepackage{amssymb}
\usepackage[accsupp]{axessibility}

\usepackage{adjustbox}
\usepackage{array}
\usepackage{booktabs}
\usepackage{colortbl}
\usepackage{float,wrapfig}
\usepackage{hhline}
\usepackage{multirow}
\usepackage{subcaption} 
\usepackage{enumitem}

\def\paperID{8} 
\def\confName{CVPR}
\def\confYear{2024}

\title{EfficientViT-SAM: Accelerated Segment Anything Model \\ Without Accuracy Loss}

\author{
Zhuoyang Zhang$^{1,3*}$, Han Cai$^{2,3*}$, Song Han$^{2,3}$ \\
$^{1}$Tsinghua University, $^{2}$MIT, $^{3}$NVIDIA
\\
\url{https://github.com/mit-han-lab/efficientvit}
}

\begin{document}
\maketitle
\let\thefootnote\relax\footnotetext{$^*$Work done during an internship at NVIDIA.}

\begin{abstract}
We present EfficientViT-SAM, a new family of accelerated segment anything models. We retain SAM's lightweight prompt encoder and mask decoder while replacing the heavy image encoder with EfficientViT. For the training, we begin with the knowledge distillation from the SAM-ViT-H image encoder to EfficientViT. Subsequently, we conduct end-to-end training on the SA-1B dataset. Benefiting from EfficientViT's efficiency and capacity, EfficientViT-SAM delivers 48.9$\times$ measured TensorRT speedup on A100 GPU over SAM-ViT-H without sacrificing performance. Our code and pre-trained models are released at \url{https://github.com/mit-han-lab/efficientvit}.
\end{abstract}
\section{Introduction}

\indent Segment Anything Model (SAM) \cite{kirillov2023segment} is a family of image segmentation models pretrained on a high-quality dataset with 11M images and 1B masks. SAM provides astounding zero-shot image segmentation performance and has many applications, including AR/VR, data annotation, interactive image editing, etc. 

Despite the strong performance, SAM is highly computation intensive, restricting its applicability in time-sensitive scenarios. In particular, SAM's main computation bottleneck is its image encoder, which requires 2973 GMACs per image at the inference time. 

To accelerate SAM, numerous efforts have been made to replace SAM's image encoder with lightweight models. For example, MobileSAM \cite{zhang2023faster} distills the knowledge of SAM's ViT-H model into a tiny vision transformer. EdgeSAM \cite{zhou2023edgesam} trains a purely CNN-based model to mimic ViT-H, employing a meticulous distillation strategy with the prompt encoder and mask decoder involved in the process. EfficientSAM \cite{xiong2023efficientsam} leverages the MAE pretraining method to improve the performance. 

\begin{figure}
\centering
\includegraphics[width=\linewidth]{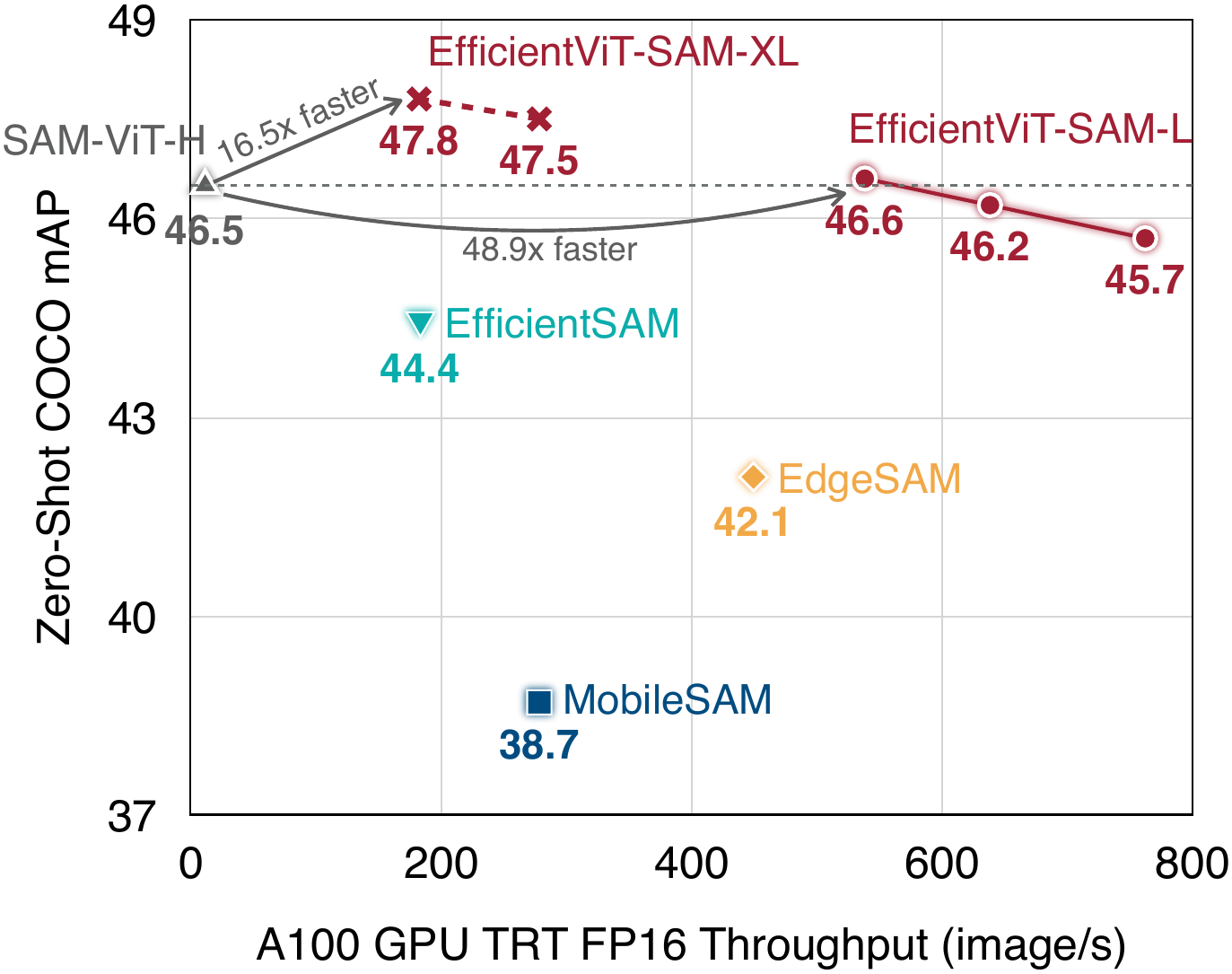}
\caption{\textbf{Throughput vs. COCO Zero-Shot Instance Segmentation mAP.} As far as we know, EfficientViT-SAM is the first accelerated SAM model that matches/outperforms SAM-ViT-H's \cite{kirillov2023segment} zero-shot performance, delivering the SOTA performance-efficiency trade-off.}
\label{fig:teaser}
\end{figure}
\begin{figure*}
\centering
\includegraphics[width=\linewidth]{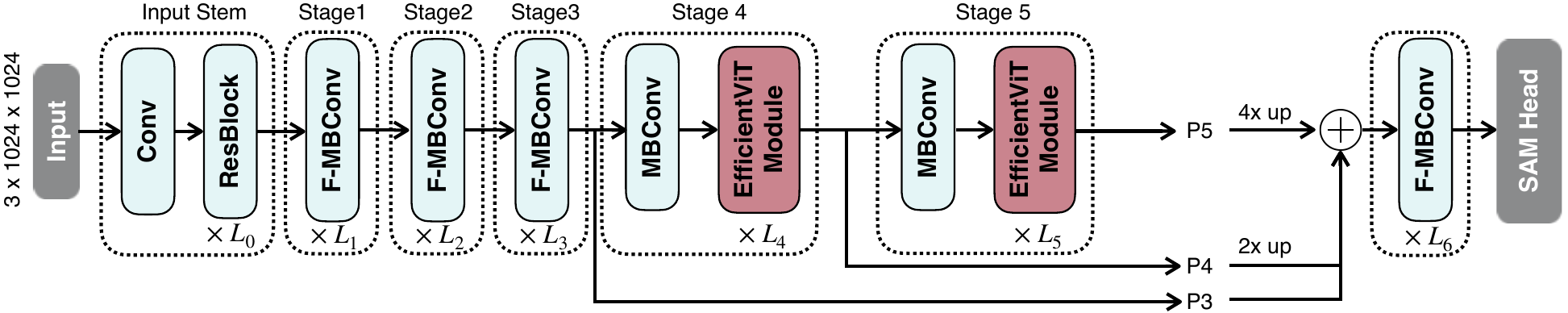}
\caption{\textbf{Macro Architecture of EfficientViT-SAM-XL.} `ResBlock' refers to the basic building block from ResNet34 \cite{he2016deep}. `F-MBConv' refers to the fused MBConv block from \cite{tan2021efficientnetv2}. `EfficientViT Module' is the building block from \cite{cai2022efficientvit}. }
\label{fig:method}
\end{figure*}

While these methods reduce the computation cost, they all suffer from significant performance drops (Figure~\ref{fig:teaser}). This work introduces \textbf{EfficientViT-SAM} to address this limitation by leveraging EfficientViT \cite{cai2022efficientvit} to replace SAM's image encoder. Meanwhile, we retain the lightweight prompt encoder and mask decoder architecture from SAM. Our training process consists of two phases. First, we train the image encoder of EfficientViT-SAM using SAM's image encoder as the teacher. Second, we train EfficientViT-SAM end-to-end on the whole SA-1B dataset \cite{kirillov2023segment}.

We thoroughly evaluate EfficientViT-SAM on a series of zero-shot benchmarks, including point-prompted segmentation, box-prompted segmentation, and in-the-wild segmentation. EfficientViT-SAM provides a significant performance/efficiency boost over all prior SAM models. In particular, on the COCO dataset \cite{lin2014microsoft}, EfficientViT-SAM achieves 48.9$\times$ higher throughput on A100 GPU without mAP drop compared with SAM-ViT-H \cite{kirillov2023segment}.

\section{Related Work}

\subsection{Segment Anything Model}
SAM \cite{kirillov2023segment} has gained widespread recognition as a milestone in the field of computer vision, showcasing its exceptional performance and generalization in image segmentation. SAM defines image segmentation as a promptable task, that aims to generate a valid segmentation mask given any segmentation prompt.
To achieve this objective, SAM utilizes an image encoder and a prompt encoder to process the image and provided prompts. The outputs from both encoders are then fed into a mask decoder, which generates the final mask prediction. SAM is trained on a large-scale segmentation dataset comprising over 11 million images with more than 1 billion high-quality masks, enabling robust zero-shot open-world segmentation. SAM has shown its high versatility in a wide range of downstream applications, including image in-painting \cite{yu2023inpaint}, object tracking \cite{cheng2023segment, yang2023track}, and 3D generation \cite{liu2023one, shi2023zero123++}. Nevertheless, the image encoder component of SAM imposes significant computational costs, leading to high latency that restricts its practicality in time-sensitive scenarios. Recent works \cite{zhang2023faster, zhou2023edgesam, xiong2023efficientsam, shu2023tinysam} have been focused on improving the efficiency of SAM, aiming to address its computational limitations.

\subsection{Efficient Deep Learning Computing}
Improving the efficiency of deep neural networks is critical when deploying them in real-world applications on both edge and cloud platforms. Our work is related to efficient model architecture design \cite{howard2017mobilenets,cai2018proxylessnas} that aims to improve the performance-efficiency trade-off by replacing inefficient model architectures with efficient ones. Our work is also related to knowledge distillation \cite{hinton2015distilling} that uses pretrained teacher models to guide the training of student models. Additionally, we can combine EfficientViT-SAM with other parallel techniques to further boost efficiency, including pruning \cite{han2015learning}, quantization \cite{han2015deep}, and hardware-aware neural architecture search \cite{cai2019once}.

\section{Method}
We propose EfficientViT-SAM, which harnesses EfficientViT \cite{cai2022efficientvit} to accelerate SAM. In particular, our approach preserves the prompt encoder and mask decoder architecture from SAM while replacing the image encoder with EfficientViT. We design two series of models, EfficientViT-SAM-L and EfficientViT-SAM-XL, offering a balanced trade-off between speed and performance. Subsequently, we train EfficientViT-SAM using the SA-1B dataset~\cite{kirillov2023segment} in an end-to-end fashion.

\begin{table*}[t]
\setlength{\tabcolsep}{5pt}
\small\centering
\begin{tabular}{lcccc}
\toprule
  & $\#$Params(M) & $\#$MACs(G) & Throughput (image/s) & COCO mAP  \\
\midrule 
SAM-ViT-H~\citep{kirillov2023segment} & 641.1 & 2973 & 11 & 46.5 \\
\midrule 
MobileSAM~\citep{zhang2023faster} & 9.8 & 39 & 278 & 38.7 \\
EdgeSAM~\citep{zhou2023edgesam} & 9.6 & 20 & 449 & 42.1 \\
EfficientSAM~\citep{xiong2023efficientsam} & 25.3 & 247 & 183 & 44.4\\
\midrule 
EfficientViT-SAM-L0 & 34.8  & 35 & 762 & 45.7 \\
EfficientViT-SAM-L1 & 47.7 & 49 & 638 & 46.2\\
EfficientViT-SAM-L2 & 61.3 & 69 & 538 & 46.6\\
\midrule 
EfficientViT-SAM-XL0 & 117.0 & 185 & 278 & 47.5 \\
EfficientViT-SAM-XL1 & 203.3 & 322 & 182 & 47.8 \\
\bottomrule
\end{tabular}
\caption{\textbf{Runtime Efficiency Comparison}. We benchmark the throughput on a single NVIDIA A100 GPU with TensorRT, fp16.}
\label{table:efficiency}
\end{table*}

\subsection{EfficientViT}
EfficientViT \cite{cai2022efficientvit} is a family of vision transformer models for efficient high-resolution dense prediction. Its core building block is a multi-scale linear attention module that enables the global receptive field and multi-scale learning with hardware-efficient operations. Specifically, it substitutes the inefficient softmax attention with lightweight ReLU linear attention to have the global receptive field. By leveraging the associative property of matrix multiplication, ReLU linear attention can reduce the computational complexity from quadratic to linear while preserving functionality. In addition, it enhances the ReLU linear attention with convolution to mitigate its limitation in local feature extraction. More details are available in the original paper \cite{cai2022efficientvit}.

\subsection{EfficientViT-SAM}

\paragraph{Model Architecture.} The macro architecture of EfficientViT-SAM-XL is demonstrated in Figure~\ref{fig:method}. Its backbone consists of five stages. Similar to EfficientViT \cite{cai2022efficientvit}, we use convolution blocks in the early stages while using EfficientViT modules in the last two stages. We fuse the features from the last three stages by upsampling and addition. The fused feature is fed to the neck comprising several fused MBConv blocks and then fed to the SAM head.

\paragraph{Training.}
To initialize the image encoder, we begin by distilling the image embedding of SAM-ViT-H into EfficientViT. We employ the L2 loss as the loss function. For the prompt encoder and mask decoder, we initialize them by loading the weights from SAM-ViT-H. Then, we train EfficientViT-SAM on the SA-1B dataset in an end-to-end manner.

In the end-to-end training phase, we randomly choose between the box prompt and the point prompt with equal probability. In the case of the point prompt, we randomly select 1-10 foreground points from the ground-truth mask to ensure our model performs effectively for various point configurations. In the case of the box prompt, we utilize the ground-truth bounding box. We resize the longest side to 512/1024 for EfficientViT-SAM-L/XL models and pad the shorter side accordingly. We select up to 64 randomly sampled masks per image. To supervise the training process, we use a linear combination of focal loss and dice loss, with a 20:1 ratio of focal loss to dice loss. Similar to the approach taken in SAM to mitigate ambiguity, we predict three masks simultaneously and only back-propagate the lowest loss. 

We train EfficientViT-SAM on the SA-1B dataset for 2 epochs, utilizing a batch size of 256. The AdamW optimizer is employed with a momentum of $\beta_1$ = 0.9 and $\beta_2$ = 0.999. The initial learning rate is set to 2e$^{-6}$/1e$^{-6}$ for EfficientViT-SAM-L/XL, which is decayed to 0 using a cosine decay learning rate schedule. Regarding data augmentation, we use the random horizontal flip. 

\section{Experiment}

\begin{table}[t]
\small\centering
\setlength{\tabcolsep}{2pt}
\scalebox{0.85}{
\begin{tabular}{l|ccc|ccc}
\toprule
 & \multicolumn{3}{c|}{COCO} & \multicolumn{3}{c}{LVIS}  \\
\midrule
 & 1 click & 3 click & 5 click & 1 click & 3 click & 5 click\\
\midrule
SAM-ViT-H~\citep{kirillov2023segment} & 58.4  & 69.6  & 71.4  & 59.2  & 66.0 & 66.8  \\
\midrule 
EfficientViT-SAM-XL1 & 59.8 & 71.3 & 75.3 & 56.6 & 67.0 & 71.7\\
\bottomrule
\end{tabular}}
\caption{\textbf{Zero-Shot Point-Prompted Segmentation Results.}}
\label{table:point}
\end{table}
\begin{table}[t]
\small\centering
\setlength{\tabcolsep}{2pt}
\scalebox{0.7}{
\begin{tabular}{l|cccc|cccc}
\toprule
 & \multicolumn{4}{c|}{COCO} & \multicolumn{4}{c}{LVIS}  \\
\midrule
 & $\rm mIoU$ & $\rm mIoU^S$ & $\rm mIoU^M$ & $\rm mIoU^L$ & $\rm mIoU$ & $\rm mIoU^S$ & $\rm mIoU^M$ & $\rm mIoU^L$ \\
\midrule
SAM-ViT-H~\citep{kirillov2023segment} &  77.4 & 72.3 & 80.4 & 81.8 & 77.0 & 70.6&  87.5 & 89.9  \\
\midrule
EfficientViT-SAM-XL1 & 79.9 & 75.8 & 82.2 & 83.8 & 79.9 & 74.4 & 88.4 & 91.6 \\
\bottomrule
\end{tabular}}
\caption{\textbf{Zero-Shot Instance Segmentation Results, Prompted with Ground Truth Bounding Box.}}
\label{table:box_gt}
\end{table}
\begin{table}[t]
\small\centering
\setlength{\tabcolsep}{2pt}
\scalebox{0.85}{
\begin{tabular}{l|cccc|cccc}
\toprule
 & \multicolumn{4}{c|}{COCO} & \multicolumn{4}{c}{LVIS}  \\
\midrule
 & $\rm mAP$ & $\rm AP^S$ & $\rm AP^M$ & $\rm AP^L$ & $\rm mAP$ & $\rm AP^S$ & $\rm AP^M$ & $\rm AP^L$ \\
\midrule
SAM-ViT-H~\citep{kirillov2023segment} &  46.5 & 30.8 & 51.0 & 61.7 & 44.2 & 31.8 & 57.1 & 65.3  \\
\midrule
MobileSAM~\citep{zhang2023faster} & 38.7  & 23.7  & 42.2 & 54.3 & 37.0 & 24.7  & 47.8 & 59.1  \\
EdgeSAM~\citep{zhou2023edgesam} & 42.1 & 26.6 & 46.7 & 56.9 & 39.8 & 28.6 & 51.3 & 59.3  \\
EfficientSAM~\citep{xiong2023efficientsam} & 44.4 & 28.4 & 48.3 & 60.1 & 41.5 & 29.7 & 53.4 & 62.2  \\ 
\midrule
EfficientViT-SAM-L0 & 45.7 & 28.2 & 49.5 & 63.4 & 41.8 & 28.8 & 53.4 & 64.7 \\
EfficientViT-SAM-L1 & 46.2 & 28.7 & 50.4 & 64.0 & 42.1 & 29.1 & 54.3 & 65.0\\
EfficientViT-SAM-L2 & 46.6 & 28.9 & 50.8 & 64.2 & 42.7 & 29.4 & 55.1 & 65.5 \\
\midrule 
EfficientViT-SAM-XL0 & 47.5 & 30.0 & 51.5 & 64.6 & 43.9 & 31.2 & 56.2 & 65.9\\
EfficientViT-SAM-XL1 & 47.8 & 30.5 & 51.8 & 64.7 & 44.4 & 31.6 & 57.0 & 66.4\\
\bottomrule
\end{tabular}}
\caption{\textbf{Zero-Shot Instance Segmentation Results, Prompted with ViTDet Boxes.}}
\label{table:box_vitdet}
\vspace{-15pt}
\end{table}

In this section, we begin by conducting a comprehensive analysis of the runtime efficiency of EfficientViT-SAM in Section \ref{exp:efficiency}. Subsequently, we evaluate the zero-shot capability of EfficientViT-SAM on the COCO ~\citep{lin2014microsoft} and LVIS ~\citep{gupta2019lvis} datasets, which were not encountered during the training process. Two distinct tasks are performed: point-prompted instance segmentation in Section \ref{exp:point} and box-prompted instance segmentation in Section \ref{exp:box}. These tasks individually assess the effectiveness of the point prompt and box prompt features of EfficientViT-SAM. We also provide results on SGinW benchmark~\cite{zou2023generalized} in Section \ref{exp:sginw}. 

\subsection{Runtime Efficiency}
\label{exp:efficiency}

We compare the model parameters, MACs, and throughput of EfficientViT-SAM with SAM and other acceleration works. Results are shown in Table \ref{table:efficiency}. We conduct the throughput measurements on a single NVIDIA A100 GPU with TensorRT optimization. Our results show that compared to SAM, we achieve an impressive acceleration of 17 to 69 times. Furthermore, despite having more parameters than other acceleration works, EfficientViT-SAM demonstrates significantly higher throughput due to its effective utilization of hardware-friendly operators.

\subsection{Zero-Shot Point-Prompted Segmentation}
\label{exp:point}

We assess the zero-shot performance of EfficientViT-SAM in segmenting objects based on point prompts in Table \ref{table:point}. We adopt the point selection method described in \citep{kirillov2023segment}. That is the initial point is selected as the point located farthest from the object boundary. Each subsequent point is chosen as the farthest point from the boundary of the error region, which is defined as the area between the ground truth and the previous prediction. The performance is reported using 1/3/5 clicks on COCO and LVIS dataset, with the mIoU (mean Intersection over Union) serving as the metric. Our results demonstrate superior performance compared to SAM, particularly when additional point prompts are provided.

\subsection{Zero-Shot Box-Prompted Segmentation}
\label{exp:box}

We evaluate the zero-shot performance of EfficientViT-SAM in object segmentation using bounding boxes. We first input ground truth bounding boxes to the model, and the results are presented in Table \ref{table:box_gt}. The mIoU(mean Intersection over Union) is reported for all objects, as well as separately for small, medium, and large objects. Our approach surpasses SAM by a significant margin on the COCO and LVIS dataset. Next, we employ an object detector, ViTDet~\citep{li2022exploring}, and utilize its output boxes as prompts for the model. The results in Table \ref{table:box_vitdet} demonstrate that EfficientViT-SAM achieves superior performance compared to SAM. Notably, even the lightest version of EfficientViT-SAM significantly outperforms other acceleration works by a large margin.

Additionally, we evaluate the performance of EfficientViT-SAM on the COCO dataset using YOLOv8~\cite{Jocher_Ultralytics_YOLO_2023} and Grounding DINO~\citep{liu2023grounding} as the object detectors. YOLOv8 is a real-time object detector suitable for real-world applications. On the other hand, Grounding DINO is capable of detecting objects using text prompts, allowing us to perform object segmentation based on textual cues. The results presented in Table \ref{table:box_more} reveal the outstanding performance of EfficientViT-SAM in comparison to SAM.\\
\begin{table}[t]
\small\centering
\setlength{\tabcolsep}{2pt}
\scalebox{0.85}{
\begin{tabular}{l | c c c c | c c c c}
\toprule
 & \multicolumn{4}{c|}{YOLOv8} & \multicolumn{4}{c}{GroundingDINO} \\
\midrule
 & $\rm mAP$ & $\rm AP^S$ & $\rm AP^M$ & $\rm AP^L$ & $\rm mAP$ & $\rm AP^S$ & $\rm AP^M$ & $\rm AP^L$ \\
\midrule
SAM-ViT-H~\citep{kirillov2023segment} & 43.8 & 26.1 & 48.1 & 60.4 & 46.9 & 31.5 & 51.8 & 64.4 \\
\midrule
EfficientViT-SAM-XL1 & 44.7 & 26.0 & 48.9 & 62.9 & 48.2 & 31.5 & 52.6 & 67.3 \\
\bottomrule
\end{tabular}}
\caption{\textbf{Zero-Shot Instance Segmentation Results on COCO, Prompted with YOLOv8/Grounding DINO Boxes.}}
\label{table:box_more}
\end{table}

\subsection{Zero-Shot In-the-Wild Segmentation}
\label{exp:sginw}
The Segmentation in the Wild benchmark~\cite{zou2023generalized} consists of 25 zero-shot in-the-wild segmentation datasets. We equip EfficientViT-SAM with Grounding DINO as box prompts and perform zero-shot segmentation. 
SAM achieves an mAP of 48.7, whereas EfficientViT-SAM achieves a higher score of 48.9.

\subsection{Qualitative Results}

\begin{figure}
\centering
\includegraphics[width=\linewidth]{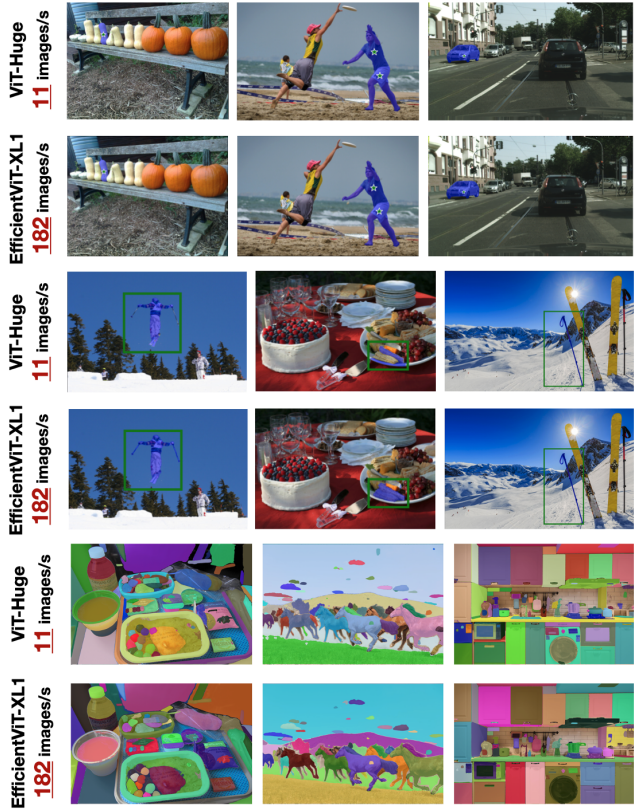}
\caption{\textbf{Qualitative Segmentation Results of EfficientViT-SAM under Point, Box, and Everything Mode.}}
\label{fig:results}
\vspace{-15pt}
\end{figure}

Figure \ref{fig:results} showcases the qualitative results of EfficientViT-SAM when provided with point prompt, box prompt, and segment-everything mode. The results demonstrate that EfficientViT-SAM excels in segmenting both large and small objects.

\section{Conclusion}
In this work, we introduced EfficientViT-SAM, which utilizes EfficientViT to replace the image encoder of SAM. EfficientViT-SAM achieved a significant efficiency boost over SAM without sacrificing performance across various zero-shot segmentation tasks. We have open-sourced our code and pre-trained models on GitHub to the community. 


{
    \small
    \bibliographystyle{unsrt}
    \bibliography{ref}
}

\end{document}